\documentclass[10pt,twocolumn,letterpaper]{article}
\PassOptionsToPackage{dvipsnames,table}{xcolor}

\usepackage{cvpr}
\usepackage[dvipsnames]{xcolor}

\newcommand{\name}[1]{DiffVQA}

\newif\ifdraft
\drafttrue
\ifdraft
\newcommand{\smc}[1]{{\color{blue}[\textbf{SM:} #1]}}

\else
\newcommand{\smc}[1]{}

\fi

\definecolor{cvprblue}{rgb}{0.21,0.49,0.74}

\usepackage{times}
\usepackage{graphicx}
\usepackage{amsmath}
\usepackage{amssymb}
\usepackage{booktabs}
\usepackage{rotating}
\usepackage{paralist}
\usepackage{caption}
\usepackage{subcaption}
\usepackage[linesnumbered,boxed]{algorithm2e}
\usepackage{arydshln}
\usepackage{array}
\usepackage{multirow}
\usepackage{ragged2e}
\usepackage{booktabs}
\usepackage{arydshln}
\usepackage{makecell}

\usepackage[dvipsnames]{xcolor}
\usepackage[accsupp]{axessibility}
\usepackage{times}
\usepackage{array}
\usepackage{ragged2e}
\usepackage{booktabs}
\usepackage{wrapfig}

\usepackage{graphicx}
\usepackage{amsmath}
\usepackage{amssymb}
\usepackage[pagebackref=true,breaklinks=true,colorlinks,bookmarks=false,citecolor=blue,linkcolor=blue]{hyperref}
\usepackage{makecell}
\usepackage[accsupp]{axessibility}

\usepackage{color, colortbl}
\definecolor{LightCyan}{rgb}{0.88,1,1}

\hypersetup{
    colorlinks=true,
    urlcolor=magenta,
    citecolor=blue,
}

\usepackage[T1]{fontenc}
\usepackage{multirow}
\usepackage{booktabs}
\usepackage{comment}
\usepackage{color}

\usepackage{textcomp}
\usepackage{siunitx}
\usepackage{tabulary}
\usepackage{makecell}
\usepackage{multirow}
\usepackage{booktabs}
\usepackage{hyperref}

\makeatletter
\newcommand\figref{Figure~\ref}
\newcommand{\tabref}[1]{Table~\ref{#1}}
\newcolumntype{P}[1]{>{\centering\arraybackslash}p{#1}}
\newcolumntype{M}[1]{>{\centering\arraybackslash}m{#1}}
\let\ts@includegraphics\includegraphics

\usepackage{float}
\usepackage{lipsum}   
\usepackage{rotating} 
\usepackage{stfloats} 
\usepackage[export]{adjustbox} 



\usepackage[capitalize]{cleveref}
\crefname{section}{Sec.}{Secs.}
\Crefname{section}{Section}{Sections}
\Crefname{table}{Table}{Tables}
\crefname{table}{Tab.}{Tabs.}

\def\paperID{9426} 
\def\confName{CVPR}
\def\confYear{2025}

\begin{document}

\title{\name{}: Video Quality Assessment Using Diffusion Feature Extractor}

\author{
Wei-Ting Chen\textsuperscript{1,3†}
\quad Yu-Jiet Vong\textsuperscript{1†}
\quad Yi-Tsung Lee\textsuperscript{1}
\quad Sy-Yen Kuo\textsuperscript{1}\\
\quad Qiang Gao\textsuperscript{2}
\quad Sizhuo Ma\textsuperscript{2*}
\quad Jian Wang\textsuperscript{2*}
\\\\
\hspace{-8mm}\textsuperscript{1}National Taiwan University\quad \textsuperscript{2}Snap Inc.\quad \textsuperscript{3}Microsoft
}

\maketitle


\begin{abstract}
Video Quality Assessment (VQA) aims to evaluate video quality based on perceptual distortions and human preferences. Despite the promising performance of existing methods using Convolutional Neural Networks (CNNs) and Vision Transformers (ViTs), they often struggle to align closely with human perceptions, particularly in diverse real-world scenarios. This challenge is exacerbated by the limited scale and diversity of available datasets. To address this limitation, we introduce a novel VQA framework, \name{}, which harnesses the robust generalization capabilities of diffusion models pre-trained on extensive datasets. Our framework adapts these models to reconstruct identical input frames through a control module. The adapted diffusion model is then used to extract semantic and distortion features from a resizing branch and a cropping branch, respectively. To enhance the model's ability to handle long-term temporal dynamics, a parallel Mamba module is introduced, which extracts temporal coherence augmented features that are merged with the diffusion features to predict the final score. Experiments across multiple datasets demonstrate \name{}'s superior performance on intra-dataset evaluations and its exceptional generalization across datasets. These results confirm that leveraging a diffusion model as a feature extractor can offer enhanced VQA performance compared to CNN and ViT backbones.
\end{abstract}

\newcommand\blfootnote[1]{%
\begingroup
\renewcommand\thefootnote{}\footnote{#1}%
\addtocounter{footnote}{-1}%
\endgroup
}

\blfootnote{† Equal contribution}
\blfootnote{* Co-corresponding authors}
\blfootnote{Project Page: \href{https://github.com/DiffVQA/DiffVQA}{https://github.com/DiffVQA/DiffVQA}}

\begin{figure}[t!]
  \centering
  \begin{subfigure}[b]{0.48\columnwidth} 
    \includegraphics[width=\textwidth]{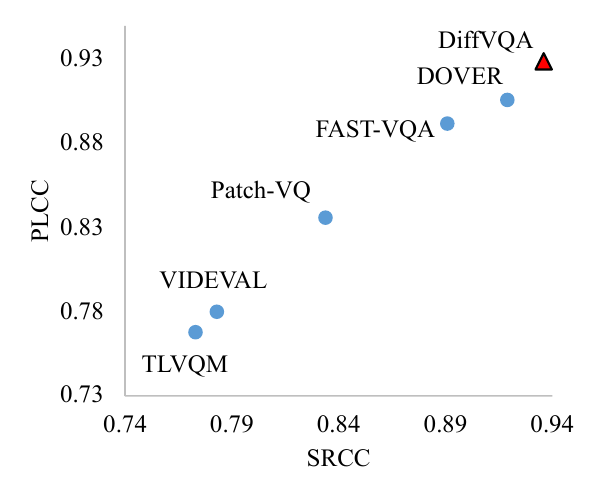} 
    \caption{Intra-dataset}
    \label{fig:subim1}
  \end{subfigure}
  \hspace{0.1cm} 
  \begin{subfigure}[b]{0.48\columnwidth} 
    \includegraphics[width=\textwidth]{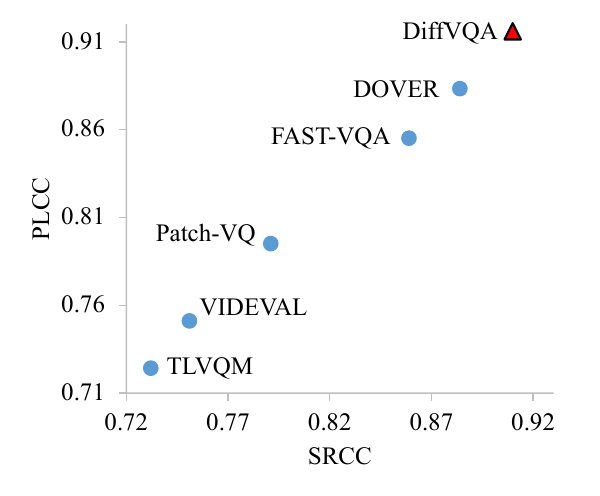} 
    \caption{Cross-dataset}
    \label{fig:subim2}
  \end{subfigure}
\caption{\textbf{PLCC and SRCC performance across intra- and cross-dataset evaluations.} \name{}, represented by red triangles, shows superior performance against established methods like DOVER \cite{wu2023exploring}, FAST-VQA \cite{wu2022fast}, TLVQM \cite{korhonen2019two}, VIDEVAL \cite{tu2021ugc}, and Patch-VQ \cite{ying2021patch}. 
We evaluate \name{} on the KoNViD-1k \cite{hosu2017konstanz} dataset using models trained on KoNViD-1k for intra-dataset evaluation and LSVQ \cite{ying2021patch} for cross-dataset evaluation, demonstrating its robust generalization capabilities across different datasets.}
  \label{fig:teasor}
\end{figure}

\section{Introduction}
In today’s digital landscape, social media platforms profoundly impact how we communicate and share life's moments, with millions of users uploading and sharing videos daily on Facebook, Instagram, TikTok, and Snapchat~\cite{wang2021rich}. Ensuring high video quality is crucial for providing a satisfying user experience, emphasizing the importance of Video Quality Assessment~\cite{tu2021rapique}. 

Current VQA methods are typically categorized into full-reference (FR) and no-reference (NR) algorithms, each with their limitations in practical scenarios. Although FR algorithms~\cite{hekstra2002pvqm} are precise, they require access to high-quality reference videos which are not always available. Conversely, NR algorithms~\cite{wang2021rich,wu2023exploring} do not rely on reference videos and are thus more versatile, but still struggle with accurately mimicking human perception. This paper explores the challenges associated with NR VQA.

NR VQA techniques are primarily categorized into prior-based and learning-based approaches. Prior-based strategies~\cite{mittal2012making,mittal2012no,tu2021ugc} often utilize handcrafted features to capture spatial and temporal aspects, yet they typically yield constrained performance due to their limited adaptability to diverse content and distortion types. In contrast, the proliferation of extensive datasets~\cite{hosu2017konstanz,lu2024kvq} has catalyzed the development of learning-based methods~\cite{wu2022disentangling,wu2022fast,gotz2021konvid}. These techniques leverage deep neural networks such as CNNs~\cite{he2016deep} and ViTs~\cite{dosovitskiy2020image}, which have demonstrated substantial promise in accuracy metrics. 
Despite these advancements, learning-based approaches still face significant challenges in accurately mirroring human perception, especially when confronted with complex real-world distortions and varied scenarios. This misalignment is further amplified by the restricted scale and diversity of datasets available.

To cope with these challenges, we develop \name{}, a novel NR VQA model that leverages generative diffusion models~\cite{sohl2015deep}. 
Pre-trained on large-scale datasets, diffusion models~\cite{rombach2022high} have revolutionized content creation by generating high-quality visual content that aligns with human perceptual preferences through sophisticated diffusion and denoising processes.
Due to their proven generalizability~\cite{chen2025unirestore,lin2024diffbir}, diffusion models have been adapted to tasks such as image segmentation \cite{saharia2022palette} and image restoration \cite{xia2023diffir}, consistently demonstrating strong performance. 
Inspired by the ability of diffusion models to generate visually pleasing content aligned with human perception while demonstrating strong generalization capabilities, this paper explores a central question: Can diffusion models also serve as powerful feature extractors to enhance VQA performance? However, leveraging diffusion models for VQA poses challenges. Pre-trained models are designed for text-to-image generation, limiting their ability to directly extract image features. Additionally, their focus on high-fidelity synthesis leaves their capacity to model low-quality visuals—key for capturing distortion cues in VQA—largely unexplored~\cite{Rombach_2022_CVPR}.

To solve these limitations, we adapt Stable Diffusion~\cite{rombach2022high} to a frame feature extractor. 
By incorporating a control module~\cite{zhang2023adding} to guide the diffusion model in generating identical reconstructions of input frames, the model functions as an effective feature extractor, capturing both essential semantic and distortion features from each frame.
Furthermore, to improve temporal dynamics and long-range dependencies in video sequences, we incorporate a Mamba module~\cite{zhu2024vision,liu2024vmamba}—an efficient mechanism known for its linear complexity in sequence processing. This module uses state space equations to enhance temporal feature extraction, improving video quality assessment by establishing comprehensive connections across frames. 
Refined features from both the diffusion and Mamba modules are merged to improve the evaluation of video quality.

We evaluate the proposed \name{} model across a diverse collection of VQA datasets to validate its efficacy. Our extensive testing includes both intra-dataset and cross-dataset scenarios, where \name{} consistently demonstrates promising performance. These results not only validate the model’s ability to extract robust features that closely align with human perceptual standards, but also highlight its excellent generalization capabilities. The contributions of this work are summarized as follows:
\begin{compactitem}
    \item We introduce \name{}, a no-reference VQA model that utilizes diffusion-based generative models for accurate video quality assessment. Our results demonstrate that the adapted diffusion model serves as an effective feature extractor for VQA.
    \item To enhance the integration of temporal features, we introduce the Mamba module to strengthen the diffusion features and further improve the performance.
\end{compactitem}

\section{Related Work}
\subsection{Diffusion Model}
Diffusion models (DMs)~\cite{sohl2015deep} are generative models that systematically corrupt data with noise and subsequently learn to reverse this corruption. They have been successfully adopted in image generation tasks \cite{ho2022cascaded}. Latent Diffusion Models (LDMs)~\cite{rombach2022high} extend the idea by operating in a latent space to reduce computational demands. 
Demonstrating remarkable versatility and robustness, DMs have excelled in a wide range of applications, from text-to-image synthesis \cite{wu2023harnessing} to image restoration \cite{xia2023diffir,chen2025unirestore}.

Research efforts are increasingly focused on enhancing the control mechanisms of DMs to enable personalization and task-specific modifications. Innovations such as prompt adjustment and CLIP feature manipulation are leading this charge \cite{avrahami2022blended,nichol2021glide}. A significant advancement in this area is ControlNet \cite{zhang2023adding}, which incorporates task-specific conditions like segmentation maps into large pre-trained models. This development has facilitated new applications such as style transfer \cite{chen2023controlstyle} and domain adaptation \cite{shen2024controluda}, showcasing the broadened utility of DMs across various domains.

\subsection{Video Quality Assessment}
VQA quantifies the perceived quality of videos as experienced by human viewers and is broadly categorized into \textit{prior-based} and \textit{learning-based} approaches.

\noindent\textbf{Prior-based VQA} employs handcrafted features to capture spatial and temporal distortions, leveraging techniques such as natural scene statistics (NSS), artifacts identification, and motion analysis. Several frameworks like NIQE~\cite{mittal2012making} and VIIDEO~\cite{mittal2015completely} utilize NSS to predict video quality. Advanced models such as TLVQM~\cite{korhonen2019two} and VIDEVAL~\cite{tu2021ugc} meticulously compute and select complex quality-aware features for more accurate assessments. However, the factors affecting video quality are quite complex and cannot be adequately captured by these handcrafted features alone, often resulting in a gap between the assessed quality and actual perceptual quality experienced by viewers.

\noindent\textbf{Learning-based VQA} employs learning technologies such as CNNs and ViT to mine deep semantic features and perform quality regression. VSFA~\cite{li2019quality} utilizes a DNN architecture enhanced with GRUs~\cite{cho2014learning}, illustrating the model’s capability in temporal analysis. Similarly, MDVSFA~\cite{li2021unified} combines features extracted using a pre-trained ResNet-50~\cite{he2016deep} with GRUs for effective temporal modeling. Additionally, other approaches involve 3D-CNN models and Inception-ResNet-V2, as in Bi-LSTM~\cite{schuster1997bidirectional} and graph convolutional networks~\cite{kipf2016semi} to enrich feature extraction capabilities. Transformer-based methods like LSCT~\cite{you2021long} process frame-wise perceptual quality features through a convolutional Transformer, enhancing the precision of quality predictions. Recent advancements such as FAST-VQA~\cite{wu2022fast} and DOVER~\cite{wu2022disentangling} introduce efficient strategies to manage and assess high-resolution videos, pushing the boundaries of VQA technology. While learning-based methods offer substantial adaptability and scalability, they are constrained by the limited size of available training datasets. This limitation becomes particularly pronounced when these models are applied to diverse real-world content and distortions, often resulting in diminished effectiveness in accurately assessing video quality under varied conditions.

\subsection{State Space Models} 
State Space Models (SSM), originally conceptualized within control theory~\cite{kalman1960new}, have been repurposed for deep learning to enhance the modeling of sequential data. These models scale linearly with sequence lengths, a capability that enables efficient processing of long sequences~\cite{gu2021combining,gu2021efficiently,smith2022simplified}. The introduction of the Structured State Space Sequence Model (S4)~\cite{gu2021efficiently} marked a notable advancement. Its further development into the S5 model~\cite{smith2022simplified}, which integrates MIMO SSM and parallel scanning techniques, has been instrumental in improving the handling of long-range dependencies. Building on this foundation, the Mamba model~\cite{gu2023mamba} introduces a selective SSM mechanism that optimizes for linear-time inference, making it highly adaptable across various applications, including natural language processing and complex visual tasks~\cite{zhu2024vision,lu2024videomambapro,he2024multi}.

\section{Proposed Method}
\subsection{Preliminary}
\noindent\textbf{Stable Diffusion} (SD) \cite{rombach2022high}
employs a pre-trained auto-encoder (VAE) to convert images from pixel space \(x\) into a compressed latent space \(z_0\) using an encoder and then reconstructs them back to pixel space using a decoder. As a result, both diffusion and denoising processes can be efficiently executed within the latent space.

The diffusion phase involves the incremental addition of Gaussian noise to the encoded latent representation \(z_0\), generating a series of progressively noisier states:
\begin{equation}
    z_t = \sqrt{\bar{\alpha}_t}z_0 + \sqrt{1 - \bar{\alpha}_t}\epsilon,
\label{eq:forward}
\end{equation}
where \(\epsilon\) is a noise vector sampled from a standard Gaussian distribution \(\mathcal{N}(0, \mathbf{I})\), $\alpha_t = 1-\beta_t$, $\bar{\alpha}_t=\Pi_{s=1}^t \alpha_s$, \(\beta_t\) represents the noise variance at time step \(t\). As \(t\) increases, \(z_t\) converges towards a Gaussian distribution.

The reverse process aims to reconstruct the original latent representation \(z_0\) from the noisy latent representation \(z_t\) by iteratively removing the added noise. 
This denoising step is achieved by a U-Net denoted by \(\epsilon_\theta\), which is designed to predict the noise component. 
The sample \(z_{t-1}\) at timestep \(t-1\) is derived from \(z_t\) as:
\begin{equation}  
z_{t-1} = \frac{1}{\sqrt{\alpha_t}} \left(z_t - \frac{1 - \alpha_t}{\sqrt{1 - \bar{\alpha}_t}} \epsilon_\theta(z_t, t, c)\right) + \sqrt{\beta_t} \epsilon_t,
\label{eq:reverse}
\end{equation}
where \(c\) can be a condition, such as text conditions or image conditions. $\epsilon_t$ is drawn from $\mathcal{N}(0, \mathbf{I})$.

The optimization goal for the denoising step \(\epsilon_\theta\) is given by the following loss function:
\begin{equation}
\scalebox{0.9}{$
    \mathcal{L}_\text{Diff} = \mathbb{E}_{z_0, c, t, \epsilon}[||\epsilon - \epsilon_{\theta}(\sqrt{\bar{\alpha}_t}z_0 + 
    \sqrt{1 - \bar{\alpha}_t}\epsilon, c, t)||_2],
$}
\end{equation}

\begin{figure}[t!]
    \centering
    \includegraphics[width=\linewidth]{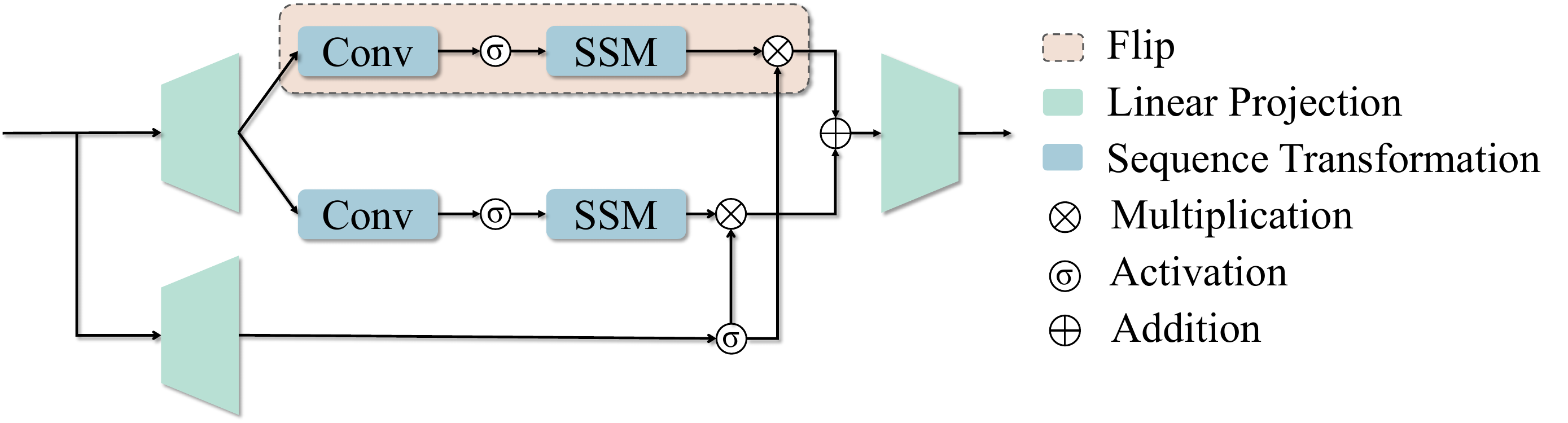}
    \caption{\textbf{Illustration of Bidirectional Mamba Block.} The initial normalization and the final residual are omitted for simplification.}
    \label{fig:mamba}
\end{figure}

\noindent\textbf{State Space Models}~\cite{kalman1960new} represent 1D functions or sequences $x(t) \in \mathbb{R} \rightarrow y(t) \in \mathbb{R}$ through a latent state $h(t) \in \mathbb{R}^N$, governed by linear ordinary differential equations (ODEs):
\begin{align}
h'(t) &= Ah(t) + Bx(t), \\
y(t) &= Ch(t) + Dx(t),
\end{align}
where $A \in \mathbb{R}^{N \times N}$ is the system evolution matrix. $B \in \mathbb{R}^{N \times 1}$, $C \in \mathbb{R}^{1 \times N}$, and $D \in \mathbb{R}$ are transformation matrices.

These continuous models are typically discretized through the zero-order hold (ZOH) method~\cite{ogata1995discrete} to facilitate practical deep-learning applications:
\begin{align}
h_k &= \overline{A} h_{k-1} + \overline{B} x_k, \\
y_k &= C h_k + D x_k,\\
\text{where}\qquad \overline{A} &= \exp(\Delta A), \\
\overline{B} &= (\Delta A)^{-1} (\exp(\Delta A) - I) \cdot \Delta B.
\end{align}

Building on the structured state-space sequence model (S4), Mamba~\cite{gu2023mamba} introduces a selective mechanism that allows system parameters $\overline{A}$, $\overline{B}$, and the timescale $\Delta$ to dynamically adapt to input data. Moreover, bidirectional adaptations for visual tasks such as Vision Mamba~\cite{zhu2024vision,liu2024vmamba} have been proposed. As shown in~\figref{fig:mamba}, such models process flattened visual sequences through simultaneous forward and backward passes, enhancing spatial awareness and processing efficiency. 
These enhancements make Mamba a versatile tool for complex image and video analysis.

\begin{figure}[t!]
    \centering
    \hspace{-0.5cm}\includegraphics[width=1.0\linewidth]{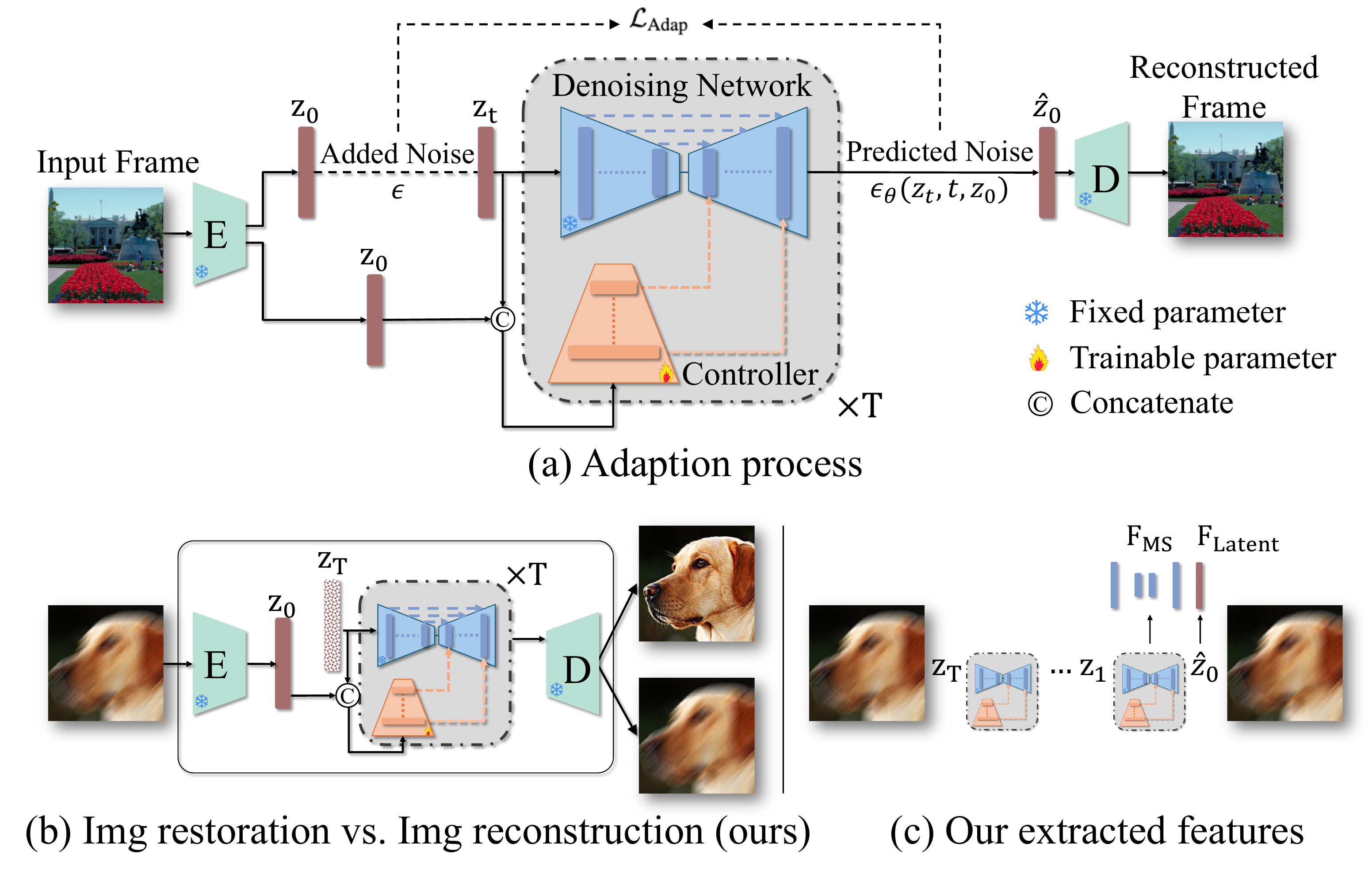}
\caption{\textbf{Adaptation process of the Diffusion Feature Extractor} shown in (a). $\mathcal{L}_{\text{Adap}}$ is computed between the added noise \( \epsilon \) and the predicted noise \( \epsilon_{\theta} \). (b) A similar architecture can be used for image restoration, but we repurpose it for image reconstruction here. (c) During inference, we use \(\hat{z}_0\), along with the features from Denoising Network at time step \( t = 0 \) as the extracted features.}
    \label{fig:diffusion_stage}
\end{figure}

\subsection{Adapting Diffusion Model to Feature Extractor}
\label{sec:adapt}
To leverage the strong generalizability of the pre-trained diffusion model to enhance VQA accuracy, two key challenges need to be tackled.
First, unlike existing pre-trained feature extractors (e.g., ViTs \cite{dosovitskiy2020image} and CLIP~\cite{radford2021learning}), pre-trained diffusion models are designed for text-to-image generation and cannot directly extract features from input images. 
Additionally, most pre-trained diffusion models are designed to generate high-quality images, so their capability to generate low-quality images~\cite{Rombach_2022_CVPR}, which is essential for capturing distortion information relevant to VQA, remains relatively unexplored. To address these challenges, we propose adapting the pre-trained diffusion model to reconstruct identical input video frames, with the help of an additional Controller~\cite{zhang2023adding}. The underlying hypothesis is: If the adapted model can indeed generate reconstructions that closely match the original inputs, it must extract the critical semantic and distortion information as its internal representation, which can be repurposed to predict the video quality.

\noindent\textbf{Adaptation Process.}
As shown in \figref{fig:diffusion_stage} (a), the process begins with encoding the input frame using the pre-trained VAE of SD, transforming it from pixel space into latent space \( z_{0} \). Next, a forward diffusion process is applied to \( z_{0} \), as described in~\eqref{eq:forward}, generating the noisy latent \( z_{t} \). This noisy latent is subsequently refined by the denoising network through the reverse diffusion process, as outlined in \eqref{eq:reverse}, gradually restoring it to the original content. Simultaneously, a Controller module takes the concatenated \( z_{t} \) and \( z_0 \) as input to provide control signals that guide the reverse diffusion process. 
The Controller consists of a trainable copy of the encoder and the middle block from the denoising network, with all the parameters initialized from the pre-trained denoising network. Features from the Controller are then merged with the decoder features of the denoising network across multiple scales via zero convolutions~\cite{zhang2023adding}. To accommodate the increased input dimensions resulting from the concatenation of \( z_{0} \) and \( z_{t} \), the first convolutional layer of the Controller is modified.

\begin{figure}[t!]
    \centering
    \includegraphics[width=1.0\linewidth]{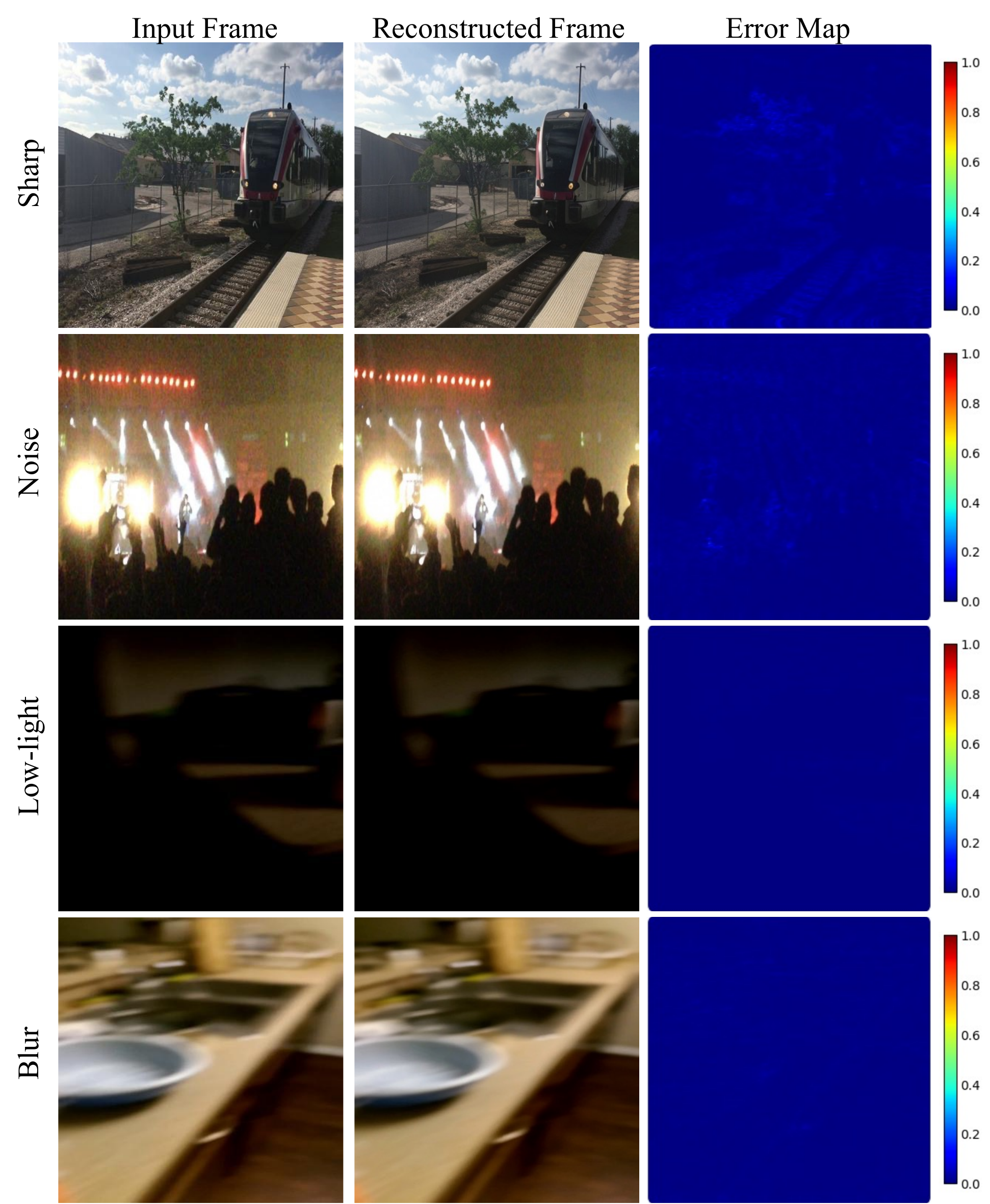}
\caption{\textbf{Examples of reconstructed results from the adapted diffusion model, along with error maps showing the differences between the input and reconstructed frames.} After adaptation, the diffusion model reconstructs the input frames with minimal error.}
    \label{fig:reconstruct}
\end{figure}

During training, only the Controller is optimized, with the text input set to null. The adaption is directed by minimizing the discrepancy between the estimated and actual noise at each time step:
\begin{equation}
\label{eq:diff_loss}
\mathcal{L}_{\text{Adap}} = \mathbb{E}_{z_0, t, \epsilon} [\| \epsilon_\theta(z_t, t, z_0) - \epsilon \|_2],
\end{equation}
where \( \epsilon_\theta(z_t, t, z_0) \) represents the predicted noise. \figref{fig:reconstruct} presents the reconstructed outputs. Our adaptation allows reconstruction of video frames containing a spectrum of degradations with minimal error. A similar architecture can be used for image restoration, but we use it for image reconstruction here (Figure \ref{fig:diffusion_stage} (b)).

\noindent\textbf{Diffusion Feature Extractor.}
After the adaptation process, we utilize the adapted diffusion model to extract features from video frames. Specifically, for each input image, we set the text input to null, encode the image into the latent feature \( z_{0} \), add noise, and initiate the reverse denoising process. Using spaced-DDPM sampling~\cite{nichol2021improved}, we begin from \( t = T \) and iterate down to \( t = 0 \) (the final step). The multi-scale features from the denoising network at \( t = 0 \), along with the reconstructed latent feature \( \hat{z}_0 \), are taken as the outputs of the diffusion feature extractor, as shown in Figure \ref{fig:diffusion_stage} (c). We refer to these two features as \( \text{F}_{\text{MS}} \) and \( \text{F}_{\text{Latent}} \) in the following section.

\begin{figure*}[t!]
    \centering
    \includegraphics[width=1.0\linewidth]{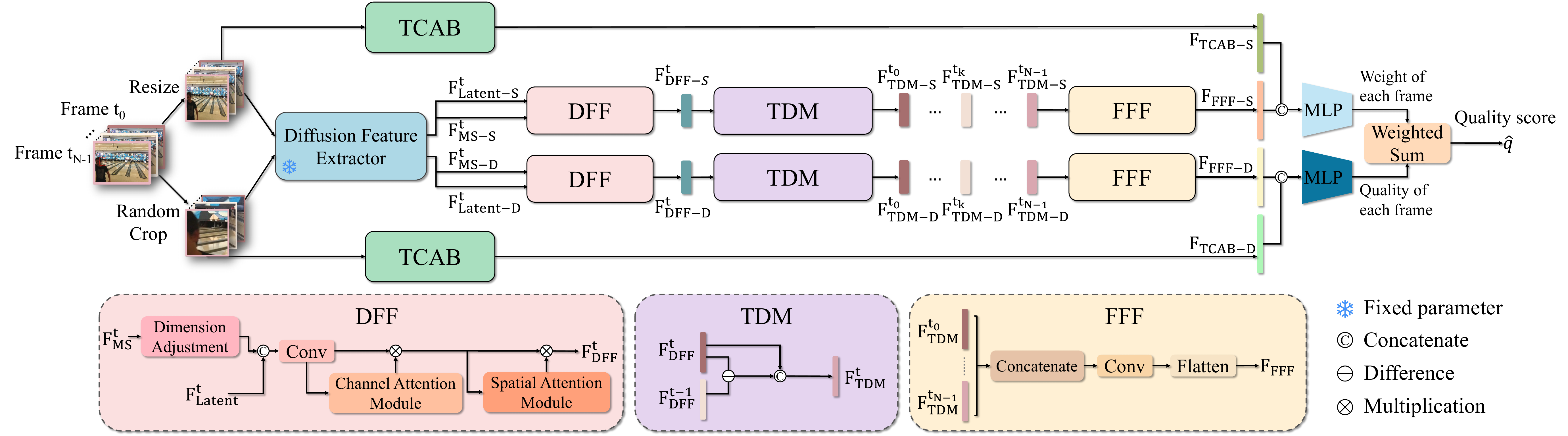}
\caption{\textbf{Architecture of \name{}.} The diffusion feature extractor extracts semantic and distortion features from video frames, which are enhanced by the DFF, TDM, and FFF modules. The TCAB are also used to capture temporal coherence in parallel. Features extracted from the resized branch are denoted with the suffix ‘‘-S’’ to represent semantic4 information, while those from the random crop branch use the suffix ‘‘-D’’ for distortion-related features.}

    \label{fig:architecture}
\end{figure*}

\begin{figure}[t!]
    \centering
    \includegraphics[width=1.0\linewidth]{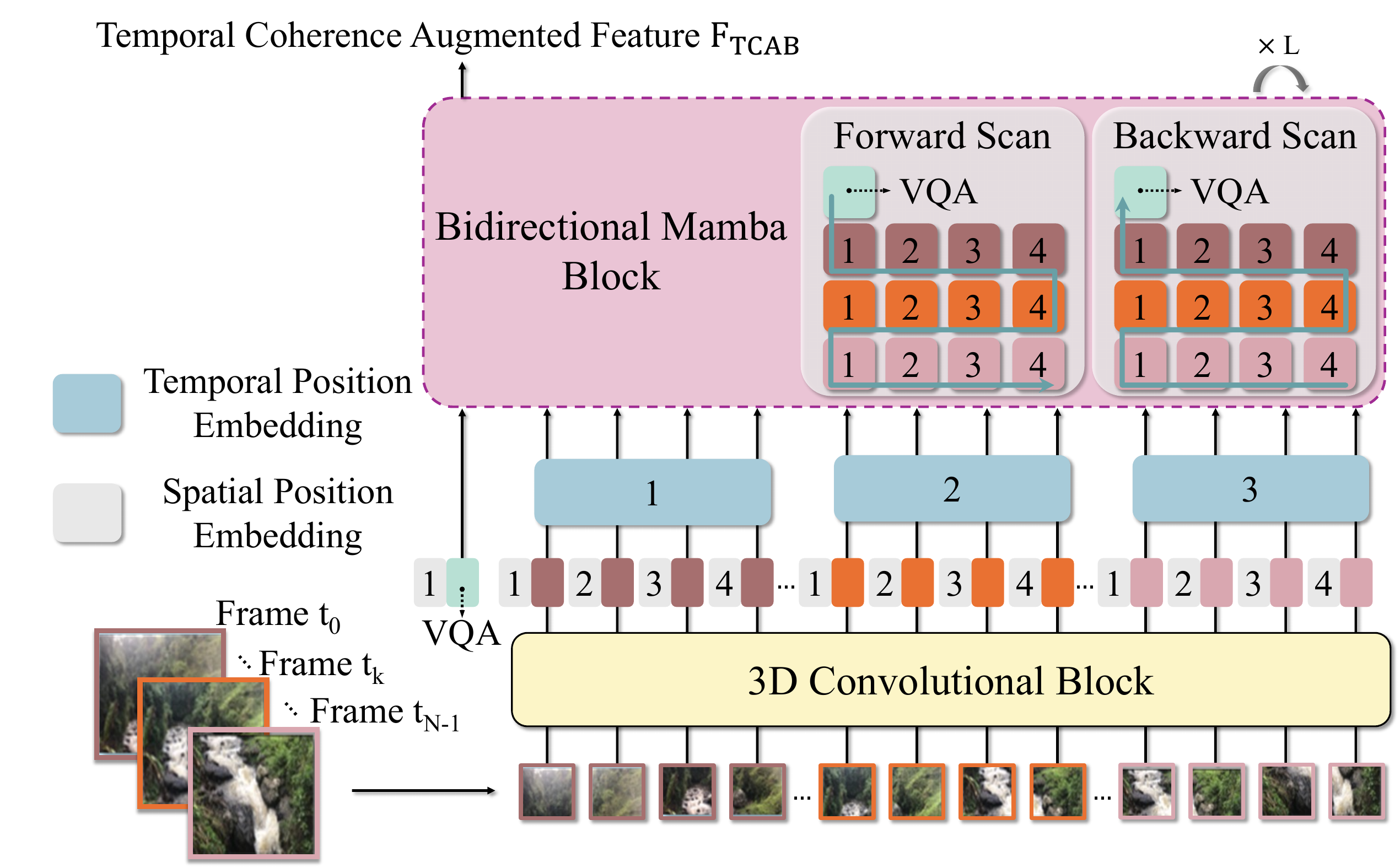}
\caption{\textbf{Architecture of the Temporal Coherence Augmentation Block.} This block utilizes a 3D convolution block to generate 3D patch embeddings. It employs a bidirectional Mamba block (\figref{fig:mamba}) to extract temporal coherence augmented features.}
    \label{fig:SSM}
\end{figure}

\subsection{Architecture of \name{}}
\noindent\textbf{Overview.} 
\figref{fig:architecture} shows the overall architecture of \name{}.
Inspired by~\cite{wu2023discovqa}, each input frame is subjected to resizing and random cropping operations, resulting in two branches passing through the diffusion feature extractor, yielding features dominantly related to semantics ($\text{F}^t_{\text{MS-S}}$ and $\text{F}^t_{\text{Latent-S}}$) and distortion ($\text{F}^t_{\text{MS-D}}$ and $\text{F}^t_{\text{Latent-D}}$) separately.
These features are integrated through the Diffusion Feature Fusion (DFF) module, producing refined diffusion features \( \text{F}^t_{\text{DFF}} \). 
Next, we apply a Temporal Difference Module to capture distortion-related temporal variations across frames, which outputs features $\text{F}^t_{\text{TDM}}$.
Then, a Frame Feature Fusion (FFF) module aggregates the \( \text{F}^t_{\text{TDM}} \) from all frames, yielding the final diffusion feature output \( \text{F}_{\text{FFF}} \).

To further strengthen temporal awareness in \name{}, we introduce a Mamba-based temporal coherence augmentation block in parallel. Finally, the temporal coherence augmented features (\( \text{F}_{\text{TCAB}} \)) and diffusion features (\( \text{F}_{\text{FFF}} \)) from the cropping branch are used to compute quality scores, while those from the resizing branch generate the weights. These features are fed into a multi-layer perceptron (MLP) to calculate the quality scores and weights for each frame to get the final video quality score \( \hat{q} \).

\noindent\textbf{Diffusion Feature Fusion (DFF).}
This module aims to fuse the output latent \( \text{F}_{\text{Latent}} \) from the denoising network and the multi-scale intermediate diffusion features \( \text{F}_{\text{MS}} \).
To this end, the multi-scale features are processed through a dimension adjustment block, which uses several convolution layers for each scale. The adjusted features are then concatenated with the latent features along the channel dimension. 
To improve the quality of the features, we apply channel attention~\cite{hu2018squeeze} and spatial attention~\cite{woo2018cbam}, resulting in the output $\text{F}^t_{\text{DFF}}$.

\noindent\textbf{Temporal Difference Module (TDM).}
Inspired by~\cite{wu2023discovqa}, this module aims to capture temporal fluctuations and subtle distortions in the video sequence. Diffusion features $\text{F}^t_{\text{DFF}}$ are combined with the residual features from the previous frame ($\text{F}^t_{\text{DFF}} - \text{F}^{t-1}_{\text{DFF}}$), resulting in refined diffusion features \((\text{F}^t_{\text{TDM}})\).

\noindent\textbf{Frame Feature Fusion (FFF).} After obtaining the refined diffusion features for all frames, we utilize the FFF module~\cite{wu2023discovqa} to combine the information from various frames \(\text{F}^{t_0}_{\text{TDM}}\) to \(\text{F}^{t_\text{N-1}}_{\text{TDM}}\). 
This module concatenates the refined diffusion features of all frames along channel dimension, and processes them through a convolution block.
The features are then flattened to form diffusion-based video features ($\text{F}_{\text{FFF}}$).

\noindent\textbf{Temporal Coherence Augmentation Block (TCAB).}
To capture comprehensive temporal correlation,
particularly for long-range sequences, we introduce the TCAB, inspired by state space models (SSM)~\cite{gu2021efficiently,smith2022simplified} and built upon the Video Mamba framework~\cite{li2024videomamba}. 

The configuration of the TCAB is depicted in \figref{fig:SSM}. The process begins by applying a 3D convolution to the video frames, denoted as $V \in \mathbb{R}^{3 \times N \times H \times W}$, where $H$, $W$, and $N$ represent the height, width, and number of frames in the video, respectively. This convolution operation transforms the video into $L$ spatiotemporal patches, denoted by $P \in R^{L \times 3}$, where $L = t \times h \times w$ with $t = N$, $h = H/16$, and $w = W/16$. These patches are subsequently transformed into a sequence of tokens $Z$, formulated as: $Z = [Z_\text{VQA
}, P] + E_{s} + E_{t}$, where $Z_\text{VQA}$ is a learnable VQA token that is prepended to the sequence to enhance the model's ability to differentiate video quality attributes. Following prior studies~\cite{arnab2021vivit}, we apply spatial position embedding $E_{s} \in R^{(hw+1) \times C}$ and temporal position embedding $E_{t} \in R^{t \times C}$ to maintain the spatiotemporal position information. The tokens $Z$ are processed through a series of stacked Bidirectional Mamba Blocks, employing a Spatial-First bidirectional scan fashion~\cite{li2024videomamba} as shown in~\figref{fig:SSM}. This module generates the temporal coherence augmented features ($\text{F}_\text{TCAB}$), enhancing the model's ability to capture dynamic changes over time.

\noindent\textbf{Final Score Prediction.} 
We concatenate \(\text{F}_{\text{FFF}}\) features, derived from both resized and randomly cropped inputs, with their respective \(\text{F}_{\text{TCAB}}\) features along the channel dimension.
These integrated features are then processed through separate MLPs. Features from the resized inputs are used to compute the weights \(w_i\) for each frame, assessing the relative importance of each frame within the sequence.
Conversely, features from the randomly cropped inputs are used to calculate the quality scores \(s_i\) for each frame, capturing the perceptual quality with a focus on distortion characteristics.
The final video quality score \(\hat{q}\) is calculated using a weighted sum of these quality scores:
$\hat{q} = \sum_{i=0}^{N-1} w_i \cdot s_i$.
%
\\\noindent\textbf{Optimization.} 
To train our \name{} network, we utilize an $\text{L}_1$ loss function to optimize the predictions:
$\mathcal{L}_{\text{VQA}} = \| \hat{q} - q \|_1,$
where \(q\) denotes the ground truth MOS.

\section{Implementation Details}
\label{sec:implementation}

\subsection{Dataset}
We evaluate our model using five UGC (User-Generated Content) VQA databases: KoNViD-1k \cite{hosu2017konstanz} (1200 videos), LIVE-VQC \cite{hosu2017konstanz} (585 videos), YouTube-UGC \cite{wang2019youtube} (1122 videos in 15 categories), LSVQ \cite{ying2021patch} (39000 videos, largest) and KVQ \cite{lu2024kvq} (4200 short-form videos).
For the LSVQ and KVQ datasets, we use the publicly available splits to evaluate our method. For KoNViD-1k, LIVE-VQC, and YouTube-UGC, we follow established protocols from previous studies~\cite{wu2023exploring, wu2022fast, ying2021patch}, dividing each dataset into training and testing subsets with an 8:2 ratio, conducting evaluations across ten different splits for each dataset and then compute the average.

\begin{table}[t!]
    \footnotesize
    \setlength\tabcolsep{3pt}
    \renewcommand\arraystretch{1.1}
    \centering
        \scalebox{0.46}{
    \resizebox{\textwidth}{!}{%
        \begin{tabular}{l|cc|cc|cc}
            \hline
            Dataset      & \multicolumn{2}{c|}{\textbf{LIVE-VQC}}   & \multicolumn{2}{c|}{\textbf{KoNViD-1k}}    & \multicolumn{2}{c}{\textbf{YouTube-UGC}}         \\ \hline
            Methods   &   SRCC$\uparrow$& PLCC$\uparrow$  & SRCC$\uparrow$& PLCC$\uparrow$         & SRCC$\uparrow$& PLCC$\uparrow$                                            \\ \hline 
            TLVQM  \cite{korhonen2019two}     & 0.799 &  0.803  & 0.773 & 0.768   & 0.669 &  0.659 \\
            VIDEVAL \cite{tu2021ugc}   & 0.752 &  0.751  & 0.783 & 0.780    & 0.779 &  0.773\\
            RAPIQUE \cite{tu2021rapique}      & 0.755 &  0.786  & 0.803 & 0.817 & 0.759 &  0.768 \\ 
            CNN+TLVQM  \cite{korhonen2020blind}  & 0.825 & 0.834 & 0.816 & 0.818  & 0.809 & 0.802  \\
            CNN+VIDEVAL \cite{tu2021ugc}   & 0.785 & 0.810 & 0.815 & 0.817  & 0.808 & 0.803 \\
            GSTVQA \cite{chen2021learning} & 0.788 &   0.796 &  0.814 &  0.825 & N/A & N/A   \\ 
            VSFA  \cite{li2019quality}       & 0.773 &  0.795  & 0.773 & 0.775   & 0.724 &  0.743   \\
            Patch-VQ  \cite{ying2021patch}   & {0.827} &  {0.837}  & 0.791 &   0.786        & N/A &  N/A \\
            CoINVQ \cite{wang2021rich} & N/A &  N/A & 0.767 &  0.764    & {0.816} &     {0.802}   \\ 
            Li \textit{et al.} \cite{li2022blindly}  & 0.834 & 0.842 & 0.834 & 0.836 & 0.818 & 0.826   \\ 
            DisCoVQA \cite{wu2023discovqa} & 0.820 & 0.826 & 0.846 &  0.849 & 0.809 & 0.808 \\ 
            SAMA \cite{liu2024scaling} & 0.860 &  0.878 & 0.892 & 0.892 & 0.881 & 0.880 \\   
            SimpleVQA \cite{sun2022deep}  &  N/A & N/A & 0.856 & 0.860 & 0.847 &  0.856  \\ 
            FAST-VQA \cite{wu2022fast}  &  0.849 & 0.862 & 0.891 & 0.892 & 0.855 & 0.852  \\ 
            MaxVQA \cite{wu2023towards} & 0.854 &  0.873 & 0.894 & 0.895 & 0.894 & 0.890 \\ 
            DOVER \cite{wu2023exploring} & 0.860 & 0.875 & 0.909 & 0.906 & 0.890 & 0.891 \\ 
            CLiF-VQA \cite{mi2024clif} & \underline{0.866} &  0.878 & 0.903 & 0.903 & 0.888 & 0.890 \\ 
            KSVQE \cite{lu2024kvq} &  0.861 & \underline{0.883} & \underline{0.922} & \underline{0.921} & \underline{0.900} & \underline{ 0.912} \\ \hline
            {\textbf{\name{}}}  &  \textbf{0.910} & \textbf{0.924} & \textbf{0.936} & \textbf{0.929} & \textbf{0.923} & \textbf{0.926}   \\ 
            \hline
        \end{tabular}}}
    \caption{\textbf{Evaluation results on multiple UGC-VQA datasets post fine-tuning.} \textbf{Bold} text highlights the best results, while \underline{underlined} text marks the second-best results.}
    \label{table:vqa_finetune}
\end{table}

\begin{table}[t!]
    \footnotesize
    \setlength\tabcolsep{3pt}
    \renewcommand\arraystretch{1.1}
    \centering
        \scalebox{1.0}{
    \resizebox{\linewidth}{!}{
        \begin{tabular}{l|cc|cc|cc|cc}
            \hline
            \multirow{2}{*}{}& \multicolumn{4}{c|}{Intra-dataset} & \multicolumn{4}{c}{Cross-dataset} \\ 
            \hline
             Test Set & \multicolumn{2}{c|}{LSVQ$_\text{test}$} & \multicolumn{2}{c|}{LSVQ$_\text{1080p}$} & \multicolumn{2}{c|}{KoNViD-1k} & \multicolumn{2}{c}{LIVE-VQC} \\ 
            \hline
            Methods & SRCC$\uparrow$ & PLCC$\uparrow$ & SRCC$\uparrow$ & PLCC$\uparrow$ & SRCC$\uparrow$ & PLCC$\uparrow$ & SRCC$\uparrow$ & PLCC$\uparrow$ \\ 
            \hline
            TLVQM \cite{korhonen2019two} & 0.772 & 0.774 & 0.589 & 0.616 & 0.732 & 0.724 & 0.670 & 0.691 \\
            VIDEVAL \cite{tu2021ugc} & 0.795 & 0.783 & 0.545 & 0.554 & 0.751 & 0.741 & 0.630 & 0.640 \\
            VSFA \cite{li2019quality} & 0.801 & 0.796 & 0.675 & 0.704 & 0.784 & 0.795 & 0.734 & 0.772 \\
             Patch-VQ$_\textit{w/o patch}$  \cite{ying2021patch} & 0.814 & 0.816 & 0.686 & 0.708 & 0.781 & 0.781 & 0.747 & 0.776 \\ 
             Patch-VQ$_\textit{w/ patch}$  \cite{ying2021patch} & 0.827 & 0.828 & 0.711 & 0.739 & 0.791 & 0.795 & 0.770 & 0.807 \\ 
            {Li \textit{et al.}  }\cite{li2022blindly} & {0.852} & {0.855} & {0.771} & {0.782} & {0.834} & {0.837} & {0.816} & {0.824} \\

            DisCoVQA \cite{wu2023discovqa} & 0.859 & 0.850 & 0.734 & 0.772 & 0.846 & 0.849 & 0.823 & 0.837 \\
            {SimpleVQA} \cite{sun2022deep} & 0.867 & 0.861 & 0.764 &  0.803 & 0.840 & 0.834 & N/A & N/A \\
            {FAST-VQA} \cite{wu2022fast} & 0.876 & 0.877 & 0.779 & 0.814 & 0.859 & 0.855 & 0.823 & 0.844 \\ 
            Q-ALIGN \cite{wu2023q} &  0.883 & 0.882 & \underline{0.797} & 0.830 & 0.865 & 0.877 & N/A & N/A \\
            CLiF-VQA \cite{mi2024clif} & 0.886 & 0.887 & 0.790 & \underline{0.832} & 0.877 & 0.874 & \underline{0.834} & \underline{0.855} \\
            DOVER \cite{wu2023exploring} & \underline{0.888} & \underline{0.889} & 0.795 & 0.830 & \underline{0.884} & \underline{0.883} & 0.832 & \underline{0.855} \\  
            \hline
            \textbf{\name{}} & \textbf{0.917} & \textbf{0.910} & \textbf{0.817} & \textbf{0.848} & \textbf{0.910} & \textbf{0.916} & \textbf{0.842} & \textbf{0.877} \\ 
            \hline
        \end{tabular}}}
    \caption{\textbf{Evaluation results on intra-dataset (LSVQ) and cross-dataset (KoNViD-1k and LIVE-VQC) using models trained on the LSVQ training set without further fine-tuning.}}
    \label{tab:vqa_all}
\end{table}

\subsection{Training and Inference Details}
\noindent\textbf{Diffusion Feature Extractor.} We use Stable Diffusion 2.1-base~\cite{Rombach_2022_CVPR} as our pre-trained backbone, with input images sized at \(384 \times 384\). The Controller is trained for 30,000 iterations using a batch size of 32 and the Adam optimizer~\cite{kingma2014adam}, configured with a weight decay of \(10^{\text{-2}}\) and a learning rate of \(10^\text{-4}\). The training is conducted on eight NVIDIA V100 GPUs using targeted VQA datasets. We adopt a Cosine noise scheduler for the diffusion process. During inference, we utilize spaced-DDPM sampling~\cite{nichol2021improved} with 10 timesteps to enhance efficiency, with a total of \( T = 1000 \) time steps.

\noindent\textbf{\name{}.} 
The VQA network is trained for 50 epochs with a batch size of 2, using the Adam optimizer with a weight decay of \(10^{-6}\) and a learning rate of \(10^{-5}\). To mitigate overfitting, a dropout rate of 10\% is applied in the linear layers. During training, the diffusion feature extractor is fixed to ensure the model focuses on optimizing the downstream VQA tasks while retaining the pre-trained feature extraction capabilities. Input frames are resized and randomly cropped to \(384 \times 384\). The entire training process is carried out on single NVIDIA V100 GPU.

\section{Experimental Results}\label{sec:exp}

\begin{table}[t!]
    \vspace{1mm}

\footnotesize
\setlength\tabcolsep{5.8pt}
\renewcommand\arraystretch{1.2}
\centering
    \scalebox{0.75}{
\resizebox{0.8\linewidth}{!}{%
\begin{tabular}{l|cc}
\toprule
Methods & SRCC$\uparrow$ & PLCC$\uparrow$ \\ \hline
VIQE ~\cite{zheng2022completely} & 0.221 & 0.397 \\
TLVQM ~\cite{korhonen2019two} & 0.490 & 0.509 \\
RAPIQUE ~\cite{tu2021rapique} & 0.740 & 0.717 \\
VIDEVAL ~\cite{tu2021ugc} & 0.369 & 0.639 \\
VSFA ~\cite{li2018has} & 0.762 & 0.765 \\
GSTVQA ~\cite{chen2021learning} & 0.786 & 0.781 \\
PVQ ~\cite{ying2021patch} & 0.794 & 0.801 \\
SimpleVQA ~\cite{sun2022deep} & 0.840 & 0.847 \\
FAST-VQA ~\cite{wu2022fast} & 0.832 & 0.834 \\
DOVER ~\cite{wu2022disentangling} & 0.833 & 0.837 \\
KSVQE ~\cite{lu2024kvq} & \underline{0.867} & \underline{0.869} \\ \hline
\textbf{\name{}} & \textbf{0.889} & \textbf{0.891} \\ \hline
\end{tabular}}}
\caption{\textbf{Evaluation results on KVQ (short-form videos).}}
\label{tab:KVQ}
\end{table}

\begin{table}[t!]
\footnotesize
    \vspace{1mm}

\setlength\tabcolsep{3pt}
\footnotesize
\centering
\scalebox{1.0}{
\resizebox{\linewidth}{!}{
\begin{tabular}{l|cc|cc|cc|cc}
\hline
Training Set & \multicolumn{4}{c|}{{LIVE-VQC}} & \multicolumn{4}{c}{{Youtube-UGC}} \\ \hline
Test Set & \multicolumn{2}{c|}{{KoNViD-1k}} & \multicolumn{2}{c|}{{Youtube-UGC}}   & \multicolumn{2}{c|}{{LIVE-VQC}} & \multicolumn{2}{c}{{KoNViD-1k}} \\ \hline
\textit{}                         & SRCC$\uparrow$                        & PLCC$\uparrow$                        & SRCC$\uparrow$                        & PLCC$\uparrow$        & SRCC$\uparrow$                        & PLCC$\uparrow$               & SRCC$\uparrow$                       & PLCC$\uparrow$                                         \\ 
\hline
TLVQM \cite{korhonen2019two} & 0.640  & 0.630& 0.218 & 0.250 & 0.488 & 0.546 & 0.556 & 0.578 \\
CNN-TLVQM \cite{korhonen2020blind}             & 0.642 & 0.631 & 0.329 & 0.367      & 0.551 & 0.578 & 0.588 & 0.619                               \\ 

VIDEVAL \cite{tu2021ugc}                    & 0.625                        & 0.621 & 0.302 & 0.318  & 0.542                         & 0.553      & 0.610 &   0.620                           \\ 
MDTVSFA \cite{li2022blindly}                     & 0.706                         & {0.711}  & {0.355} &  {0.388}      & {0.582}                        & {0.603}   & {0.649} &   {0.646}                     \\ 
BVQA \cite{li2021unified} & {0.738} & {0.721}       & {0.602}     & {0.602} &  {0.689}     & {0.727} & {0.785} & {0.782}                       \\
DisCoVQA \cite{wu2023discovqa}           & {0.792}               & {0.785}  & {0.409} & 0.432   & 0.661               & 0.685      &  0.686 &   0.697               \\ 
 MaxVQA~\cite{wu2023exploring} &  \underline{0.833} & \underline{0.831} & \underline{0.846} & \underline{0.824} & \underline{0.804} & \underline{0.812} & \underline{0.855} & \underline{0.852}\\ \hline
 \textbf{\name{}} & \textbf{0.870} & \textbf{0.871} & \textbf{0.866} & \textbf{0.842} & \textbf{0.816} & \textbf{0.835} & \textbf{0.871} & \textbf{0.881}\\  
\hline
\end{tabular}}}
\caption{\textbf{Comparative analysis of cross-dataset generalization performance.}}
\label{tab:crossval}

\end{table}

\begin{table}[t!]
    \footnotesize
    \setlength\tabcolsep{3pt}
    \renewcommand\arraystretch{1.1}
    \centering
        \scalebox{1.0}{
    \resizebox{\linewidth}{!}{
        \begin{tabular}{l|cc|cc|cc}
            \hline
            Test Set & \multicolumn{2}{c|}{LIVE-VQC} & \multicolumn{2}{c|}{KoNViD-1k} & \multicolumn{2}{c}{Performance Gap} \\ 
            \hline
            & SRCC$\uparrow$ & PLCC$\uparrow$ & SRCC$\uparrow$ & PLCC$\uparrow$ & $\Delta_{\text{SRCC}}$$\downarrow$ & $\Delta_{\text{PLCC}}$$\downarrow$ \\ 
            \hline
            ResNet-50  & 0.778 & 0.780 & 0.691 & 0.698 & 0.087 & 0.082 \\
            ResNet-101  & 0.783 & 0.787 & 0.701 & 0.706 & 0.082 & 0.081 \\
            Vision Mamba & 0.768 & 0.770 & 0.677 & 0.682 & 0.091 & 0.088 \\
            VAE & 0.786 & 0.791 & 0.705 & 0.711 & 0.081 & 0.080 \\     
            VAE* & 0.792 & 0.795 & 0.710 & 0.716 & 0.080 & 0.079 \\    
            CLIP & 0.791 & 0.795 & 0.711 & 0.720 & 0.080 & 0.075 \\
            ViT-B/16 & 0.831 & 0.818 & 0.752 & 0.740 & 0.079 & 0.078 \\
            ViT-L/16  & \underline{0.857} & \underline{0.849} & \underline{0.780} & \underline{0.772} & \underline{0.077} & \underline{0.077} \\
            \hline
            Diffusion Feature Extractor & \textbf{0.910} & \textbf{0.924} & \textbf{0.870} & \textbf{0.871} & \textbf{0.040} & \textbf{0.053} \\
            \hline
        \end{tabular}}}
\caption{\textbf{Comparative analysis of different feature extractor backbones in \name{}.} The model is finetuned on LIVE-VQC, and a lower ''$\Delta$'' value, calculated as LIVE-VQC (intra-dataset) minus KoNViD-1k (cross-dataset), indicates better generalization ability.}
\label{tab:extractor}
\end{table}

\begin{table}[t!]
\footnotesize
\setlength\tabcolsep{5.8pt}
\renewcommand\arraystretch{1.}
\centering
    \scalebox{0.5}{
\resizebox{1\linewidth}{!}{%
\begin{tabular}{l|cc}
\toprule
Metric & SRCC$\uparrow$ & PLCC$\uparrow$ \\ \hline
w/o TCAB & 0.893 & 0.907 \\
w. TCAB-S  & 0.900 & 0.908 \\
w. TCAB-D & \underline{0.905} & \underline{0.911} \\ 
w. 3D Conv & 0.895  & 0.905 \\
\hline
\textbf{\name{}} & \textbf{0.910} & \textbf{0.924} \\ \hline
\end{tabular}}}
\caption{\textbf{Effectiveness of the Temporal Coherence Augmentation Block in \name{}.} We train and test on LIVE-VQC dataset.}
\label{tab:TCAB}
\end{table}

\subsection{Performance Evaluation}
We evaluate the performance of \name{} against a diverse set of VQA methods, including VIQE ~\cite{zheng2022completely}, NIQE  \cite{mittal2012making}, TPQI \cite{liao2022exploring},
SAQI  \cite{wu2023bvqi}, TLVQM \cite{korhonen2019two}, VIDEVAL \cite{tu2021ugc}, and RAPIQUE \cite{tu2021rapique}, TLVQM \cite{korhonen2020blind}, VIDEVAL \cite{tu2021ugc}, VSFA \cite{li2019quality}, Patch-VQ \cite{ying2021patch}, CoINVQ \cite{wang2021rich}, Li \textit{et al.}~\cite{li2022blindly}, FAST-VQA \cite{wu2022fast}, DOVER \cite{wu2023exploring}, VIQE \cite{zheng2022completely}, GSTVQA \cite{chen2021learning}, PVQ \cite{ying2021patch}, SimpleVQA \cite{sun2022deep}, KSVQE \cite{lu2024kvq}, BVQA \cite{li2021unified}, DisCoVQA \cite{wu2023discovqa}, CLiF-VQA \cite{mi2024clif}, Q-ALIGN \cite{wu2023q}, and MaxVQA \cite{wu2023towards}.

\noindent\textbf{Evaluation Protocol.}
We follow common practices of existing works and employ Spearman's Rank Order Correlation Coefficient (SRCC) and Pearson's Linear Correlation Coefficient (PLCC) for quantitative evaluation.

\noindent\textbf{Evaluation on UGC-VQA Datasets.} We evaluate \name{} across multiple UGC-VQA datasets by first pre-training the model on the LSVQ dataset~\cite{ying2021patch} and then fine-tuning it on three smaller datasets: LIVE-VQC~\cite{hosu2017konstanz}, KoNViD-1k~\cite{hosu2017konstanz}, and YouTube-UGC~\cite{wang2019youtube}. As shown in \tabref{table:vqa_finetune}, \name{} demonstrates consistent results, achieving higher performance than previous VQA models on all three datasets. On the LSVQ test set and the 1080p subset, as indicated in ~\tabref{tab:vqa_all}, \name{} achieves improvements over state-of-the-art methods.

\noindent\textbf{Evaluation on Short-form UGC-VQA Dataset.} Short-form user-generated content (S-UGC) platforms engage billions of users daily. Videos on these platforms frequently encounter quality challenges due to non-professional production techniques and varied creative styles, complicating video quality assessment. To evaluate the performance on short-form videos, we use the KVQ dataset \cite{lu2024kvq}. The results, summarized in~\tabref{tab:KVQ}, show that \name{} outperforms existing VQA methods.

\noindent\textbf{Generalization Ability of \name{}.} We assess \name{}'s performance in zero-shot scenarios through cross-dataset evaluations across multiple configurations. As shown in \tabref{tab:vqa_all} and \tabref{tab:crossval}, the cross-dataset results indicate notable generalization capabilities. These findings suggest that \name{} demonstrates adaptability across various VQA datasets, highlighting the effectiveness of using generative priors from a diffusion model for video quality assessment.

\begin{figure}[t!]
  \centering
  \setlength{\tabcolsep}{1pt} 
  \scalebox{0.7}{

\begin{tabular}{cccc}
      & ViT &\hspace{0.1cm} Diffusion &  \\
      \rotatebox[origin=c]{90}{Semantic} & 
      \includegraphics[valign=m,width=.5\linewidth, height=2cm]{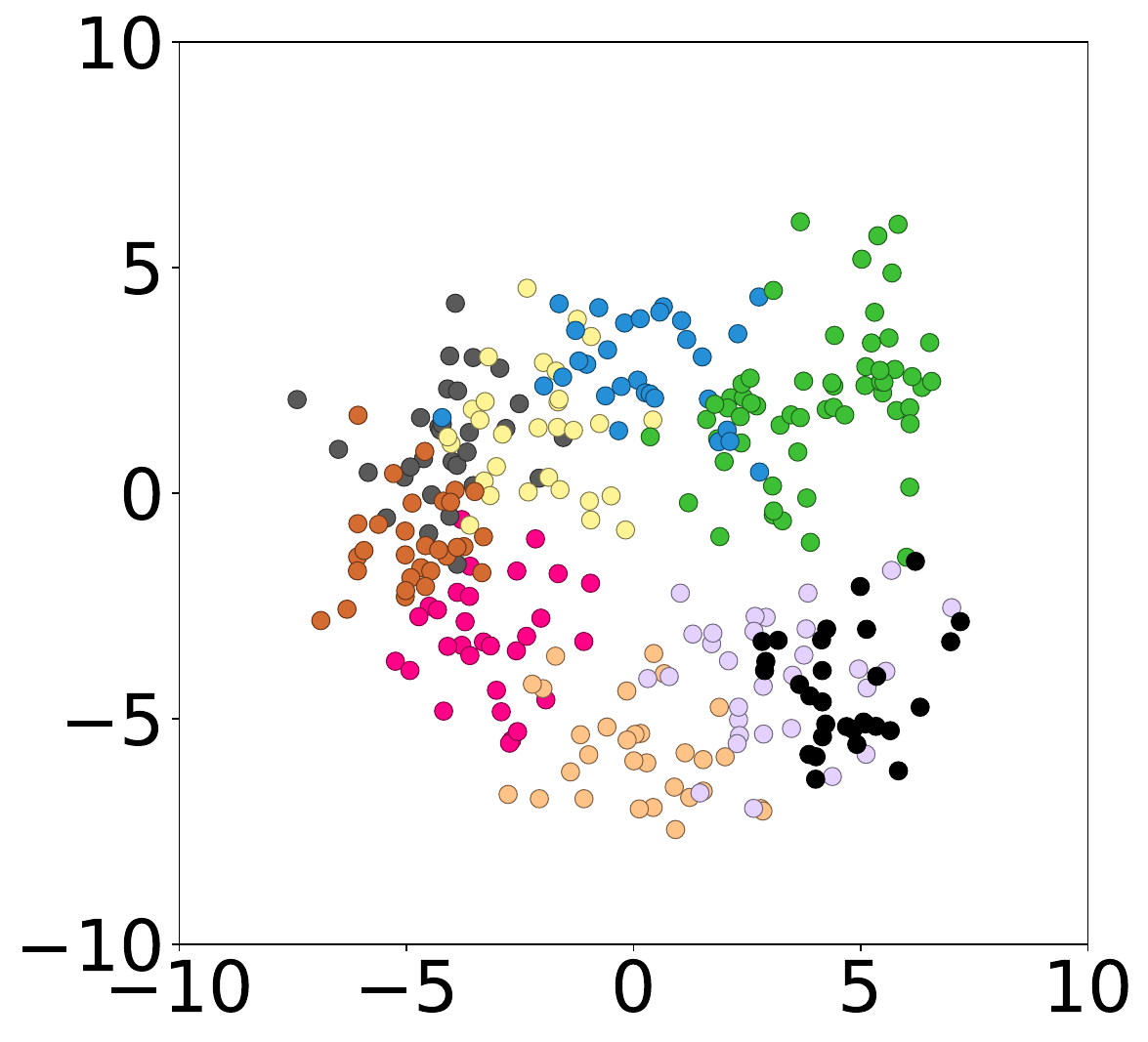} &
      \hspace{-0.2cm}
      \includegraphics[valign=m,width=.5\linewidth, height=2cm]{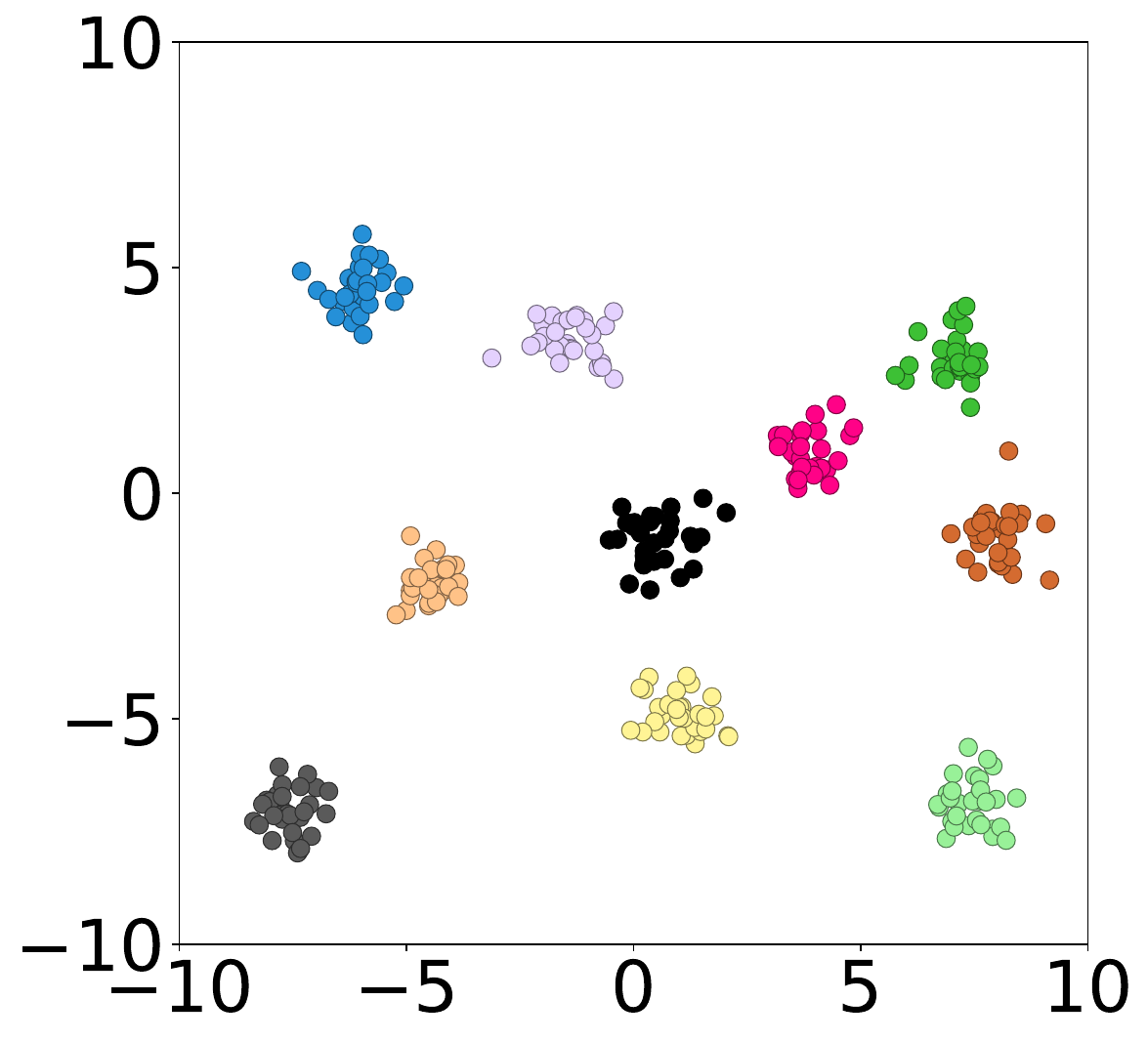} & 
      \hspace{-0.7cm}
      \raisebox{-1.6cm}{\includegraphics[width=.18\linewidth]{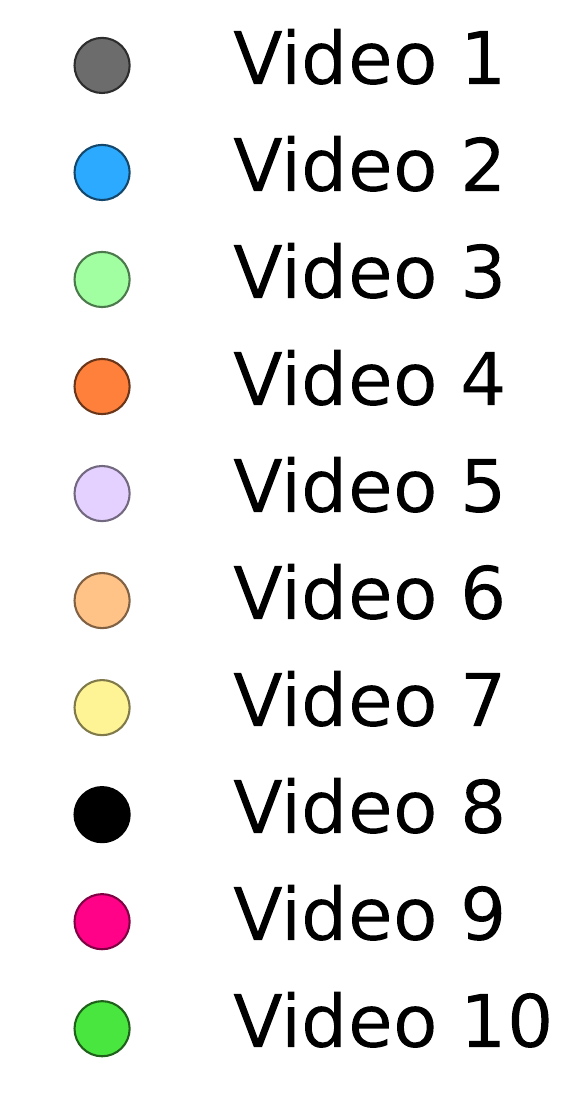}}\\
      \multicolumn{4}{c}{\vspace{-10pt}} \\ 
      
      \rotatebox[origin=c]{90}{Distortion} &
      \includegraphics[valign=m,width=.5\linewidth, height=2cm]{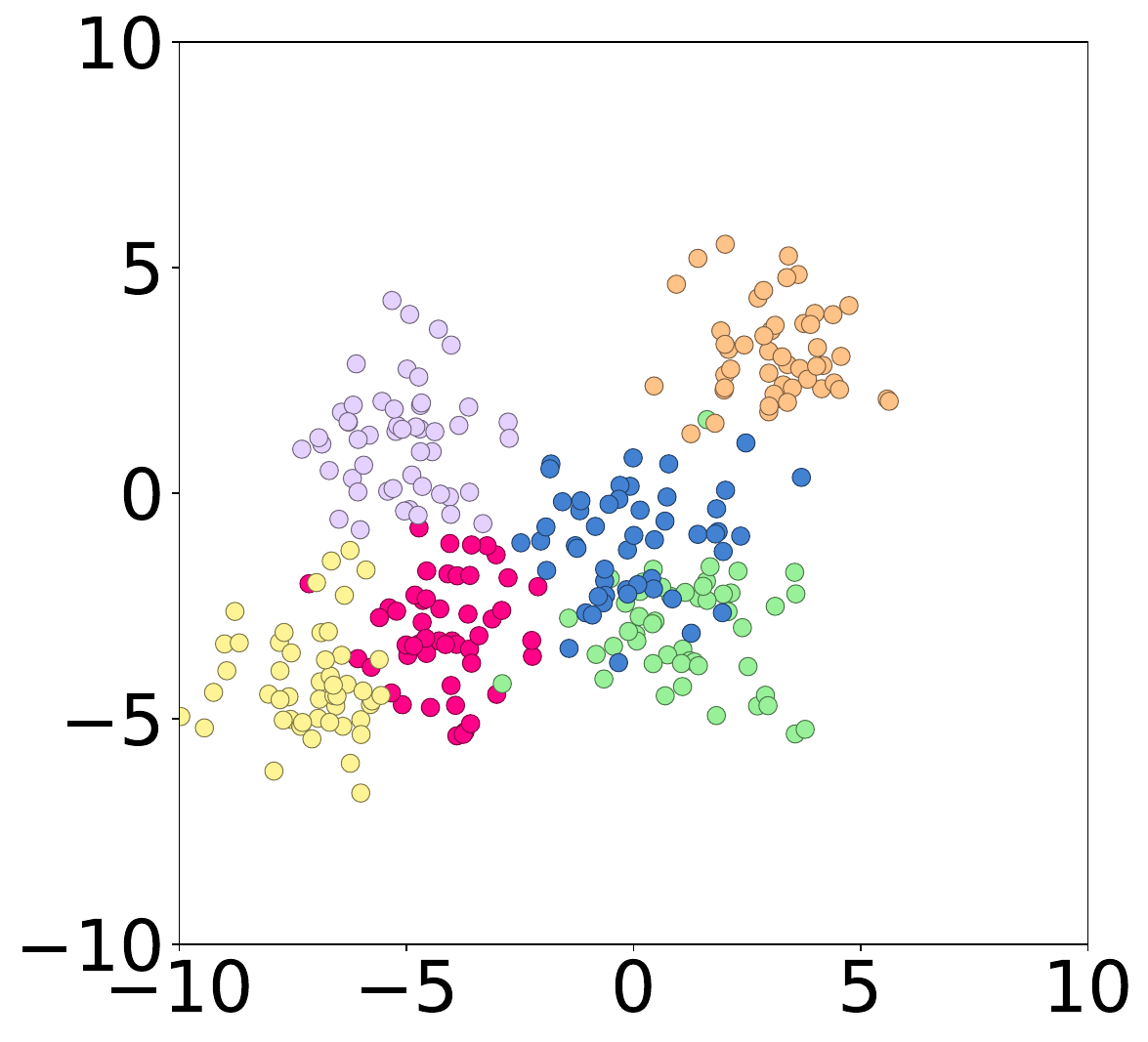} &
      \hspace{-0.2cm}
      \includegraphics[valign=m,width=.5\linewidth, height=2cm]{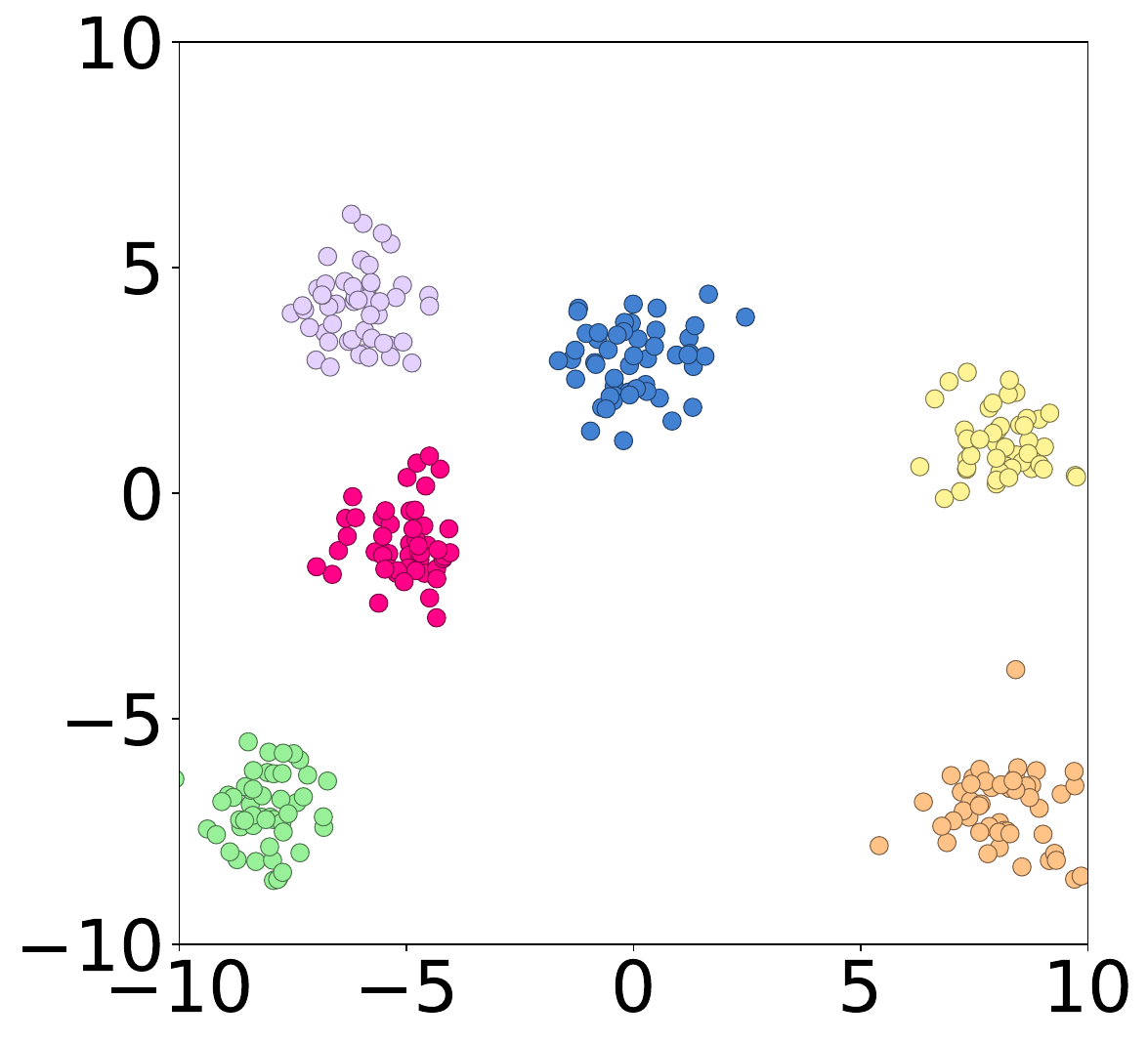} &
      \hspace{-0.2 cm}
      \raisebox{-1.6cm}{\includegraphics[width=.25\linewidth, height=1.7cm]{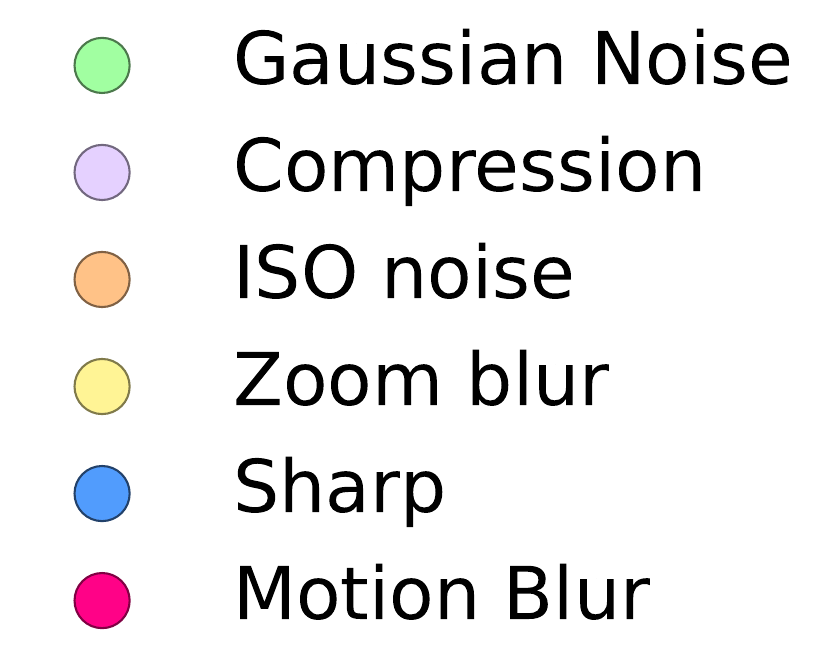}}\\
\end{tabular}
    }
\caption{\textbf{t-SNE visualization of semantic and distortion representations by ViT-L/16 and the diffusion feature extractor.}}
  \label{fig:tsne}
\end{figure}

\subsection{Ablation Study}

\noindent\textbf{Effectiveness of Different Feature Extractors.} To evaluate the influence of various feature extractors in the \name{} model, we replaced the diffusion feature extractor with CNNs (ResNet-50 and ResNet-101) \cite{he2016deep}, Vision Transformers (ViT-B/16, ViT-L/16) \cite{dosovitskiy2020image}, Vision Mamba \cite{zhu2024vision,liu2024vmamba}, CLIP~\cite{radford2021learning}, and VAE of Stable Diffusion to assess their impact on VQA performance. 
Each backbone is initialized with pre-trained weights, and feature dimension alignment is applied to their outputs to ensure compatibility with the model architecture. 
All variants are trained under identical conditions, allowing any observed performance differences to be attributed solely to feature extraction capabilities. 
Since the pre-trained VAE can be fine-tuned to reconstruct the identical input image (as described in Section \ref{sec:adapt}), we include another variant VAE* which first finetunes the VAE in this manner and then trains the entire VQA network together.
We also conduct cross-dataset evaluations on KoNViD-1k to assess generalization ability. As shown in \tabref{tab:extractor}, the diffusion feature extractor consistently demonstrates superior performance across both intra-dataset and cross-dataset scenarios,  with reduced performance gaps between them, highlighting its advantages in capturing fine features
while offering strong generalization for VQA.

We further analyze the effectiveness of different backbone architectures for extracting decoupled semantics/distortion features.
Using the Inter-4K dataset~\cite{stergiou2022adapool}, a dataset of high-quality videos, we synthesize data through two methods: First, we randomly select a frame from each of 10 videos and apply a degradation synthesis process~\cite{hendrycks2019benchmarking} with 10 types of degradation (\textit{e.g.,} blur, noise, low light, adverse weather), each with three levels of severity (\textit{i.e.,} high, medium, and low). This results in 30 frames with varying degradations per original frame, and we use the resized branch to assess the effectiveness of semantic feature extraction. Second, we introduce six types of degradation to a randomly selected frame from each of the 50 videos. This dataset, processed with random cropping, is then used to evaluate the effectiveness of distortion feature extraction. Both datasets are then processed through the ViT and Diffusion feature extractors trained in~\tabref{tab:extractor}. 
We obtain the features $\text{F}_{\text{DFF-S}}$ and $\text{F}_{\text{DFF-D}}$. We analyze and visualize them with t-SNE.
As shown in~\figref{fig:tsne}, the diffusion feature extractor yields clearly separated clusters for different semantic and distortion representations, demonstrating better capability to extract decoupled features compared to ViT.

\noindent\textbf{Effectiveness of Temporal Coherence Augmentation Block.}
To evaluate the impact of TCAB, we conduct several experimental configurations: removing TCAB from \name{} (w/o TCAB), applying TCAB only to the semantic or distortion branches (w. TCAB-S and w. TCAB-D), and substituting the Mamba module in TCAB with a 3D convolution (w. 3D Conv). These setups are compared with our proposed approach, \name{}. As shown in \tabref{tab:TCAB}, incorporating TCAB into both branches provides the highest performance by delivering comprehensive temporal information to each branch. Additionally, the Mamba module outperforms the 3D convolution, as it captures long-term temporal dependencies more effectively.

\section{Conclusion}
In this paper, we introduce a novel VQA framework, \name{}, which leverages the extensive knowledge embedded in a pre-trained diffusion model to extract features from video frames.
To further enhance temporal information extraction, we incorporate a Mamba module based on a state-space model.
Extensive experiments show that \name{} performs effectively in both intra-dataset evaluations and cross-dataset scenarios, demonstrating strong generalization abilities on unseen datasets. The results indicate that using a diffusion model as a feature extractor surpasses existing backbone feature extractors, highlighting its effectiveness in capturing essential video quality attributes.
\clearpage





{\small
\bibliographystyle{ieee_fullname}
\bibliography{main}
}

\end{document}